# Predictors of Loneliness in Older Adults Using Multimodal Analysis of Speech and Language

Vinmay Khandode[1], Sai Karthik Kosuri[1], Neil K. R. Sehgal[1,2], Adam Greene[3], Elif Alpoge[3], Elana Duffy[3], Matthew Lee Smith[4], Thomas K.M. Cudjoe[5], Sharath Chandra Guntuku[1,2]

**Affiliations:**
[1]Computer and Information Science Department, University of Pennsylvania, Philadelphia, PA, USA
[2]Leonard Davis Institute of Health Economics, University of Pennsylvania, Philadelphia, PA, USA
[3]Klaatch, a division of SeniorsTogether, Inc., New York, NY, USA
[4]Department of Health Behavior School of Public Health, Texas A&M University 212 Adriance Lab Road, #360 A, TAMU 1266 College Station, TX 77843
[5]Department of Medicine, Division of Geriatric Medicine and Gerontology, Johns Hopkins School of Medicine, Baltimore, Maryland 21224

**Corresponding Author:** Neil K. R. Sehgal, neilsehgal99@gmail.com

## Abstract:
Loneliness is a critical public health issue among older adults, linked to higher risks of depression, cognitive decline, and mortality. Scalable, objective methods for its detection remain limited, particularly in natural conversational contexts. We analyzed speech and language markers of loneliness in 310 older adults using semi-structured telephone interviews to help understand how they process feeling lonely and how their language differs at different levels of feeling loneliness. Our multimodal framework combined linguistic features (psycholinguistic dictionaries, n-grams, and topic models) with acoustic features (pitch, tone, loudness) to examine associations with self-reported loneliness scores. Both predefined and data-driven methods captured patterns in verbal content and vocal delivery. Higher loneliness was associated with negations($r = 0.11$), negative tone($r = 0.12$), and conflict-related language. Lower loneliness was linked to social references($r = -0.18$), motivational drives($r = -0.11$), and emotional richness in speech($r = -0.12$). We also found that the multimodal model ($r = 0.298$) outperforms the text-only and audio-only models. Findings suggest that loneliness manifests through both linguistic and acoustic cues, supporting the potential of speech-based analysis in psychological assessments and as an early indicator of emotional loneliness when used alongside existing assessments, rather than as standalone diagnostic tools.

## Introduction:

Loneliness is a pervasive and growing public health issue, especially among older adults [1]. As people age, they often experience significant life changes, such as retirement, bereavement, reduced mobility, and relocation, that can diminish social networks and increase the risk of social isolation [2]. Research has shown that chronic loneliness can lead to a wide range of negative health outcomes, including depression, anxiety, cardiovascular disease, cognitive decline, and even premature death [3, 4]. For instance, one meta-analysis found that loneliness increases the risk of mortality by 26% [5], underscoring its seriousness as a health risk factor on par with obesity and smoking. As a result, there is an urgent need for more effective and scalable tools to identify and mitigate loneliness in aging populations.

While several efforts have been made to proactively measure loneliness, existing approaches face significant limitations. Most assessments rely on self-reported survey instruments or screening questionnaires [6]. Although informative, these tools may not fully capture the complexity and individual variability of loneliness as expressed in natural communication. They often reduce a deeply personal and multifaceted experience into a set of predefined numerical scores, potentially overlooking the subtle linguistic signals that accompany feelings of social disconnection [7]. In particular, little is known about how older adults talk about feeling lonely and how their language differs when they report feeling lonely versus when they report feeling emotionally connected. The nuanced differences in language—whether lexical, semantic, or emotional—may provide insight into patterns that have yet to be fully understood. This study seeks to fill that gap by analyzing the content and speech pattern of semi-structured telephone interviews with older adults, focusing specifically on the linguistic characteristics that distinguish self-reported loneliness from social connectedness. Leveraging tools from natural language processing, audio signal analysis, and statistical modeling, we aim to uncover patterns in how loneliness is verbally expressed. This approach may offer an alternative perspective that moves beyond standardized scales, potentially providing richer insight into the lived experiences of older adults and laying the foundation for more personalized and linguistically-informed interventions.

Prior research has demonstrated that loneliness, social isolation, and related mental health outcomes in older adults can be reflected in both linguistic and vocal expression. Studies using linguistic analysis have shown that features such as first-person pronouns, reduced social references, and affective language are associated with loneliness, social isolation, and depression [8,9]. Complementary work using speech analysis has identified acoustic correlates of loneliness, including changes in harmonic structure, resonance, and prosodic variability, suggesting that vocal delivery may encode emotional and social states beyond lexical content alone [10,11]. Beyond speech and text, researchers have explored the use of passive sensing technologies—including ambient sensors, smart-home systems, and smartphones—to infer loneliness and social engagement among older adults, highlighting the promise of unobtrusive monitoring approaches [12–14].

More recent work has emphasized the importance of sociocultural and contextual factors in how loneliness is experienced and expressed. Qualitative studies of Black older adults describe loneliness as emerging through narratives of being misunderstood, cultural displacement, and lived experience rather than explicit emotion labeling [15]. Population-based studies among older migrants further demonstrate that language proficiency, neighborhood ethnic density, and social integration shape loneliness risk [16], while research on untreated hearing loss links auditory limitations to increased loneliness and emotional distress [17]. Longitudinal reviews show that loneliness is dynamic over time and influenced by health status, social networks, and functional changes in later life [18–20].

While some studies have combined linguistic and acoustic features through multimodal modeling, primarily for related outcomes such as depression, fewer have focused specifically on emotional loneliness in older adults using naturalistic conversational data collected longitudinally in real-world care settings [21,22]. This gap motivates the present study.

This study investigates whether linguistic and acoustic features from semi-structured telephone interviews can predict self-reported loneliness among older adults. We hypothesize that (i) both linguistic and acoustic features will show measurable associations with emotional loneliness, (ii) linguistic features reflecting negative tone and self-focus will correlate with higher loneliness, while social and affective language will correlate with lower loneliness, and (iii) a multimodal model combining both modalities will outperform unimodal approaches.

## Results:

The following analyses are exploratory and descriptive in nature and are intended to characterize patterns of association within demographic strata rather than to support causal or generalizable claims.

### *Linguistic Correlates of Loneliness*

*Full Sample Analysis*

Across all participants, LIWC categories like social referent ($r = -0.18$, $p < 0.001$), social processes ($r = -0.12$, $p < 0.001$), third-person pronouns ($r = -0.14$, $p < 0.001$), time orientation ($r = -0.07$, $p < 0.01$), and drive for affiliation ($r = -0.11$, $p < 0.001$) were associated with low loneliness. Conversely, increased negative tone ($r = 0.12$, $p < 0.001$), negative emotion ($r = 0.11$, $p < 0.001$) and cognition showing uncertainty ($r = 0.15$, $p < 0.001$) and discrepancy ($r = 0.13$, $p < 0.001$) were associated with high loneliness (Table S1). We observed similar patterns among LDA topics, e.g., social processes and third-person pronoun topics were strongly correlated to low loneliness and topics showing uncertainty strongly correlated to high loneliness. (Fig 1, Table 2).

*Gender-Based Subgroup Patterns*

Among men, we found low loneliness was highly associated with social referents ($r = -0.16$, $p = 0.03$) and talk about the future ($r = -0.15$, $p = 0.05$), whereas high loneliness was associated with cognition ($r = 0.16$, $p = 0.03$), and showing uncertainty ($r = 0.23$, $p < 0.001$) in their speech.

Among women, low loneliness was reflected in the use of social processes ($r = -0.17$, $p < 0.01$), religion ($r = -0.17$, $p < 0.01$), and references to the past ($r = -0.19$, $p < 0.001$). Conversely, high loneliness was characterized by cognition (r = 0.15, p <0.01), showing uncertainty (r = 0.19, p <0.001) and discrepancy (r = 0.15, p = 0.02) along with self-addressing pronouns (r = 0.21, p <0.001). (Table S2, Table S5)

_Racial Subgroup Results_

For White participants, low loneliness was significantly correlated with social referents (r = -0.18, p <0.01). In contrast, high loneliness was associated with self addressing speech (r = 0.18, p <0.01), uncertainty (r = 0.17, p <0.01) and discrepancy (r = 0.17, p <0.01), along with negative tone (r = 0.24, p <0.001) and negative emotions (r = 0.19, p < 0.01).

For Black participants, low loneliness was reflected by focus on the past (r = -0.19, p <0.01) and with third-person pronouns (r = -0.19, p <0.01) (Table 6). For high loneliness, significant LIWC correlation did not emerge for black participants, but two LDA topics emerged: language revolving around Africa (r = 0.21, p <0.001), and lifestyle (r = 0.25, p <0.001). The emergence of these topics should not be interpreted as indicators of loneliness per se. Rather, these topics reflect thematic variation in narrative content, capturing culturally situated experiences and conversational contexts that co-occur with self-reported loneliness in this sample. (Table S3, Table S4, Table S6)

***Acoustic Correlates of Loneliness :***

_Full Sample Analysis_

Across the entire dataset, high loneliness was associated with strong vowel sounds (r = 0.22, p<0.001) and sharper tones (r = 0.166, p<0.001). In contrast, low loneliness was associated with features such as low emotional peaks (r = –0.23, p<0.001) and monotones (r = -0.22, p<0.001) (Table 3, Table S7).

_Gender Subgroups Results_

For men, high loneliness was associated with dynamic speech (r = 0.32, p<0.001), shift in resonance (r = 0.31, p<0.001), and expressive speech (r = 0.24, p<0.001). Meanwhile, low loneliness was associated with dynamic tone (r = -0.26, p<0.001), low emotional peaks, and calmer voice (r = -0.34, p<0.001).
For women, high loneliness was characterized by harmonic richness (r = 0.29, p<0.001) and resonant shifts (r = 0.22, p<0.001), while low loneliness in women correlated to dynamic tones (r = -0.24, p<0.001) and low emotional peaks (r = -0.27, p<0.001). (Table S8)

_Racial Subgroups Results_

For White people, high loneliness was indicated by dynamic speech, strong vowel sounds (r = 0.34, p<0.001), sharp tones (r = 0.27, p<0.001), and expressive speech (r = 0.26, p<0.001), while low loneliness was indicated by less dynamic speech, weak vowel sounds (r = -0.33, p<0.001), monotone in speech (r = -0.25, p<0.001) and change in timber (r = -0.25, p<0.001).

For Black people, high loneliness was associated with loudness (r = 0.23, p<0.001), strong vowel sounds (r = 0.22, p<0.001), and harmonic richness (r = 0.19, p<0.001), while low loneliness was associated with less distinct harmonics (r = -0.23, p<0.001) and low emotional peaks and calmer voice (r = -0.28, p<0.001). (Table S9)

***Predictive Modeling of Loneliness***

*Full Dataset*

Multimodal models (combining text and audio) outperformed unimodal models in predicting CEL scores, achieving the highest Pearson correlation (r = 0.298). Among individual modalities, text-based models led by LIWC (r = 0.269) showed superior predictive strength compared to audio models like Librosa (r = 0.178) and OpenSMILE (r = 0.173). This underscores the value of integrating modalities for a more robust predictive framework.

*Gender Subgroups Results*

For men, text features (r = 0.234) surpassed audio features (r = 0.141) in predicting loneliness. The 1-to-3-gram model led among text inputs (r = 0.274), highlighting the significance of word patterns. Audio contributed minimally, as the multimodal model (r = 0.23) indicated that the predictions were primarily driven by text features.

For females, audio multimodal (r = 0.224) and text features multimodal (r = 0.229) showed comparable predictive performance. The multimodal model (r = 0.228) offered no substantial advantage, indicating that both modalities contributed similarly to loneliness prediction in this subgroup.

*Racial Subgroups Results*

For the white subgroup, audio multimodal (r = 0.286) outperformed text multimodal (r = 0.169) in predicting loneliness, with OpenSMILE (r = 0.299) and LIWC (r = 0.245) leading their respective modalities. The multimodal model achieved the highest correlation (r = 0.343), highlighting the advantage of combining audio and text for stronger predictive accuracy.

For the black subgroup, audio features slightly outperformed text in predicting loneliness, with OpenSMILE leading (r = 0.238) the combined audio model (r = 0.19). Text features showed weaker performance, while the multimodal model reached r = 0.219, offering modest gains through integration (Table 4).

Importantly, the feature representations used in this study are not novel inventions. They align with those employed in prior computational work on loneliness and related mental health outcomes. Linguistic features such as LIWC categories capturing pronoun use, affect, social processes, cognition, and negation, as well as n-gram representations, have been widely used in prior loneliness detection studies [8,9,11]. Similarly, our use of open-vocabulary topic modeling parallels prior semantic approaches that aim to capture thematic structure in loneliness narratives [8]. On the acoustic side, prosodic and spectral features derived from Librosa and OpenSMILE (e.g., pitch, loudness, formant-related measures) closely correspond to those used in prior speech-based loneliness and mental health assessments [10,17,22].

By evaluating these established linguistic and acoustic feature families jointly within a single modeling and validation framework, the present study enables a direct comparison of modalities and demonstrates that their integration yields modest but consistent performance gains over unimodal approaches. Thus, the contribution of this work lies not in proposing a new state-of-the-art model or feature set, but in empirically characterizing how previously validated representations behave when combined in a multimodal analysis of naturalistic, longitudinal interview data from older adults.

To disentangle the contribution of feature representation and modality combination, we compared text-only, audio-only, and multimodal models under an identical modeling and validation framework. Among text-based approaches, LIWC features achieved the strongest performance ($r = 0.269$), followed by n-grams ($r = 0.258$), while topic-model features (LDA) showed more modest predictive strength ($r = 0.183$). In contrast, individual acoustic feature sets such as OpenSMILE ($r = 0.173$) and Librosa ($r = 0.179$) yielded lower correlations, indicating that paralinguistic cues alone capture a smaller but meaningful portion of loneliness-related variance.

Importantly, combining complementary feature representations improved predictive performance. The combined text model (LIWC + LDA + n-grams) achieved $r = 0.283$, outperforming any individual text feature class. Similarly, combining acoustic representations improved performance relative to individual audio feature sets (combined audio $r = 0.195$). The highest performance was observed for the multimodal model integrating both text and audio features ($r = 0.298$), demonstrating that linguistic content and vocal delivery encode partially non-overlapping information related to emotional loneliness.

Subgroup analyses further highlight this complementarity. For White participants, acoustic features were particularly informative (OpenSMILE $r = 0.299$), and the multimodal model achieved a substantially higher correlation ($r = 0.343$) than either modality alone. In contrast, for male participants, text features dominated predictive performance, with minimal incremental benefit from audio features. These results indicate that the relative utility of modalities varies across demographic groups, underscoring the importance of evaluating multimodal approaches rather than relying on a single representation. (Table 4)

Because the same regression model and validation strategy were used across all experiments, performance differences reflect differences in feature representations and their combination rather than differences in model architecture.

Prior computational studies of loneliness in older adults have reported modest effect sizes when predicting loneliness from speech, language, or behavioral signals, particularly when outcomes are continuous and data are collected in naturalistic settings. Language-based studies using structured interviews or prompted responses have demonstrated statistically significant but moderate associations ($AUC \approx 0.75$ and $F1 \approx 0.73$) between linguistic features and loneliness or social isolation [8], while speech-based approaches using structured daily-life questions have reported loneliness score estimation with $R^2 = 0.57$ and high-loneliness classification accuracy of 95.6% [10]. Unobtrusive sensing studies (using smartphones, smart-home devices) conducted in real-world environments similarly report limited variance ($R^2 = 0.35$; correlation $\approx 0.48$) [12].

Consistent with this body of work, our multimodal model achieved a Pearson correlation of r = 0.298 when predicting continuous emotional loneliness scores from semi-structured wellness interviews. Importantly, our study differs from of the prior literature by jointly evaluating linguistic and acoustic features within naturalistic conversational context and under a unified modeling framework, allowing us to isolate the contribution of feature representations rather than model architecture. These findings align with prior evidence that loneliness is a diffuse and context-dependent construct, for which behavioral and linguistic signals yield reliable but inherently modest associations.

Because prior studies differ substantially in outcome formulation (binary vs. continuous), elicitation structure, and evaluation metrics (e.g., AUC, accuracy, F1 score), direct numerical comparisons should be interpreted cautiously.

## Discussion:

This study aimed to uncover how older adults express loneliness through both linguistic content and vocal characteristics in semi-structured telephone interviews. This work is not intended to propose a clinically deployable screening tool or to claim state-of-the-art prediction accuracy. Rather, it provides an empirical analysis of how loneliness manifests in naturalistic speech and language, and demonstrates that interpretable multimodal features capture complementary aspects of this experience. Our findings reveal that multiple audio and text-based feature sets are correlated with the self-reported loneliness of the interviewees, indicating that language and speech patterns carry meaningful signals associated with loneliness and may serve as an indicators of their mental well-being, and can be an integral part of the intervention infrastructure. In all the predictions performed, predictions on the text data gave better results compared to the audio data over the entire corpus. In the text data predictions, LIWC gave the most predictive results across all datasets, whereas OpenSMILE was the most predictive for audio. The integration of audio and text features yielded the highest predictive correlations in several cases and performed comparably well in others, highlighting the multimodal model's enhanced ability to capture the linear relationship between predicted and actual loneliness scores. This suggests the complementary nature of linguistic and paralinguistic cues that may help reveal the emotional and social state of older adults.

The findings of this study builds on prior work examining linguistic, acoustic, and contextual markers of loneliness in older adults. Consistent with earlier linguistic studies linking self-focus and social language to loneliness and social isolation [8,9], our results show that language-based features, particularly those captured by psycholinguistic dictionaries, are associated with self-reported emotional loneliness at the population level. At the same time, our acoustic findings align with prior speech-based research demonstrating that vocal characteristics such as harmonic structure and prosodic variation can reflect underlying emotional and social states [10,11]. Subgroup analyses revealed heterogeneity in the relative utility of linguistic and acoustic features, particularly among Black participants, for whom dictionary-based linguistic features (LIWC) showed limited associations with higher loneliness, while acoustic and open-vocabulary representations were more informative. This pattern likely reflects known limitations of predefined lexicons in capturing culturally specific modes of emotional expression rather than substantive differences in loneliness itself. Prior computational work demonstrates that linguistic markers of psychological distress often vary by race and generalize poorly across groups [23], while qualitative studies document that loneliness among Black older adults is frequently expressed through narratives of lived experience, cultural displacement, and being misunderstood rather than explicit emotion terms [15]. Importantly, topic labels (e.g., "Africa") reflect thematic word co-occurrence and should not be interpreted as indicators of loneliness per se. In contrast, open-vocabulary topic models and acoustic features may be more sensitive to broader expressive patterns that are less constrained by lexical assumptions. Our findings also resonate with population-based and longitudinal studies emphasizing the role of language, sensory capacity, and social context in shaping loneliness risk over time [16–20]. Together, these results complement prior multimodal studies by demonstrating that linguistic and acoustic features provide overlapping yet distinct signals of loneliness, supporting the view that multimodal approaches can enrich the characterization of emotional loneliness in aging populations [21,22].

These results highlight the potential for speech and language features to be explored for incorporation into tools that monitor loneliness risk. In practical terms, such systems might be integrated into telehealth workflows or senior care services to support the early identification of social withdrawal. For instance, automated analysis of conversational data during wellness check-ins could alert care coordinators to changes in speech or language patterns that may signal increasing loneliness. Importantly, this approach could reduce reliance on self-report surveys, which may be affected by stigma or recall bias. Embedding such analytic systems within existing communication channels such as call systems, virtual check-ins, or voice assistants may enable more proactive, personalized support while minimizing participant burden. At the same time, this research contributes conceptually by demonstrating that loneliness is reflected not only in the semantic content of language but also in paralinguistic aspects of communication. The complementary predictive value of audio and text features suggests that emotional and cognitive aspects of loneliness may surface differently across modalities. Linguistic markers may capture reflective or cognitive processing of social disconnection, while vocal cues such as tone or pitch variation may reflect underlying affective states. Together, these findings provide a multimodal framework for understanding loneliness expression among older adults. Importantly, the present findings should be interpreted as evidence of association rather than as support for a fully automated or clinically diagnostic system. The modest variance explained by the models reflects the inherently subjective and multifaceted nature of loneliness, which is influenced by contextual, relational, and situational factors beyond what can be captured in speech alone. Accordingly, the proposed approach is best understood as a supportive, early-signal framework that may complement existing screening tools or clinician judgment, rather than replace them. In practice, such models could be used to flag potential changes in risk over time or to prompt follow-up assessment, but not to make determinations of loneliness in isolation.

This study has several limitations. First, although the dataset spans four years and offers a longitudinal perspective, the exact timing and frequency of interviews per participant were not uniform, which may affect the resolution of temporal trends. At the same time, interview frequency varied by participant based on operational needs and participant preferences. Participants with more frequent check-ins may differ systematically from those interviewed less often in ways that correlate with both loneliness trajectories and speech patterns (e.g., greater social engagement with community staff, differential health status). This potential selection effect should be considered when interpreting longitudinal trends and generalizability. Further analysis using repeated measures per individual, or a time-series design, along with standardized interview schedules would enable stronger causal inference about how loneliness expression evolves over time. Second, while the study focuses on older adults—a group underrepresented in computational linguistics—generalizability may still be limited to populations with similar demographics, language use, and communication patterns. Third, the distribution of CEL scores was skewed, with relatively few high-loneliness responses, resulting in pronounced skewness and kurtosis. Finally, CEL is derived from self-reported questions, which, although widely used, may be influenced by stigma, memory biases, or varying thresholds for what constitutes “feeling lonely”. Future work is needed to evaluate how such speech-based indicators perform when embedded within real-world workflows and combined with clinical, social, and self-report data.

In conclusion, this study suggests that loneliness—a deeply personal and often stigmatized emotional experience—may be detected through the combined analysis of speech and language in naturalistic interviews with older adults. By leveraging a rich, longitudinal dataset and applying a diverse suite of audio and textual feature extraction methods, we demonstrated that both acoustic cues and linguistic content are meaningfully correlated with self-reported emotional loneliness. Most notably, our multimodal model achieved incremental gains consistent with feature complementarity compared to unimodal systems, reinforcing the idea that emotional states are best understood through human expression. These findings support the potential of using technology to unobtrusively detect loneliness and may lay the groundwork for integrating such tools into real-world care settings. Embedding speech-based loneliness detection into telehealth services, senior care outreach programs, and community wellness initiatives may help improve how loneliness is identified and addressed, enabling earlier intervention, personalization of support, and continuous tracking of outcomes. As populations age, the burden of loneliness, if unaddressed, grows more widespread. It is increasingly important for the health systems to consider adopting proactive, data-driven strategies to foster connection and emotional well-being. This study represents an initial step in that direction, offering a scientifically grounded framework that may inform approaches to addressing loneliness at scale.

**Methods:**

This study primarily examined emotional loneliness, assessed through the Campaign-to-End Loneliness (CEL) score [24]. This score is derived from a structured self-report question embedded within each semi-structured telephone interview conducted with older adult participants. During the interview, respondents were asked conversational, open-ended questions to obtain narrative samples, and then asked to self-report their sense of emotional connection and perceived loneliness. Their responses were systematically coded by Klaatch, a small business working to reduce loneliness for older adults, into a numerical score reflecting the degree of emotional loneliness based on participants’ degree of agreement with the question/statement. The CEL score captures a subjective, person-centered assessment of how emotionally connected or isolated participants feel. For all modeling purposes, this score was treated as a continuous variable, allowing us to assess the strength of association between various linguistic and acoustic features and the degree of reported loneliness.

***Ethics***
This study was deemed exempt from review by the University of Pennsylvania Institutional Review Board. All research was performed in accordance with the Declaration of Helsinki.

***Klaatch's Role***
Klaatch's involvement was strictly limited to: (1) developing and providing the secure telephone system and interview technology platform, (2) creating standardized training materials for community staff, (3) providing technical support for the recording system, and (4) de-identifying all data prior to transfer for research analysis. Klaatch staff had no role in participant recruitment, decisions about which residents to interview, conducting the interviews themselves, or determining interview frequency. All participant interactions were managed by the onsite SCs employed by the senior housing communities.

***Participant Consent and Eligibility***
All participants voluntarily agreed to participate in the interviews conducted as part of routine wellness check-ins within their residential communities. No inducements or payments were offered. Participants could decline any question or recording at any time and provided verbal consent prior to each interview for audio recording and secondary analysis. Eligibility was determined solely by the ability to engage in a sensible conversation—that is, to hear, understand, and respond appropriately to general conversational prompts—based on the professional judgment of onsite SCs and their supervisors. Individuals were excluded only if cognitive, hearing, or mental health impairments significantly interfered with basic conversational ability.

***Data Collection Personnel and Training***
All interviews were conducted by Service Coordinators (SCs) licensed social work professionals employed directly by the HUD Section 202 senior housing communities where participants resided. Klaatch, a small business working to reduce loneliness for older adults, provided the telephone system technology, interview protocols, and standardized training materials, but did not employ the interviewers or supervise data collection.

Each SC met federal HUD education and training standards, which include a bachelor's degree (or equivalent experience) in social work, psychology, gerontology, or a related human services field, two to three years of experience working with older adults, and completion of at least 36 hours of formal training within 12 months of hire, followed by 12 hours of annual continuing education.

For this study, SCs underwent supplementary training in conversational interviewing techniques and administration of the Campaign to End Loneliness (CEL) items. This training was administered using standardized materials developed by Klaatch in consultation with a licensed clinical social worker (training protocol provided as Supplementary Material S1). Supervisors within each community reviewed a subset of recordings for quality assurance and adherence to conversational and ethical standards. The SCs were already known to participants through their ongoing community roles, minimizing the need for additional rapport-building and ensuring natural, consistent conversational engagement.

***Data Security and Independence***

Data were fully de-identified prior to transfer for research: all names, locations, and organization identifiers were removed. Only aggregate and de-identified feature data were analyzed. The research team at the University of Pennsylvania had no financial relationship with Klaatch. All data were stored on the University of Pennsylvania's encrypted servers. All statistical analyses were conducted independently by university researchers. Klaatch staff contributed only to technology provision, training material development, and de-identification; they had no influence on analytic design, statistical methods, interpretation, or reporting of results.

***Data***

This study analyzes semi-structured interview data collected through routine wellness check-ins in senior housing communities. These interviews were conducted over the phone using Klaatch's secure telephone system, which integrates with a client portal for recording and managing data. All interviews were conducted by SCs, trained onsite staff within the HUD Section 202 senior housing communities where participants resided. Each SC met federal HUD education and training standards, which include a bachelor's degree (or equivalent experience) in social work, psychology, gerontology, or a related human services field, two to three years of experience working with older adults, and completion of at least 36 hours of formal training within 12 months of hire, followed by 12 hours of annual continuing education.

For this project, SCs underwent supplementary training in conversational interviewing techniques, administered by a licensed clinical social worker to ensure standardized administration of open-ended prompts to elicit natural, conversational speech while ensuring participant comfort and consent and the Campaign to End Loneliness (CEL) items. Each session began with verbal consent for participation and audio recording, followed by one open-ended conversational prompt selected from a predefined set (e.g., daily activities, meaningful memories, hobbies, or community engagement) to encourage several minutes of continuous speech. Interviewers were trained to use active listening, neutral follow-up prompts, and minimal interruption to maintain naturalistic dialogue. After the conversational segment, participants completed the Campaign to End Loneliness (CEL) self-report items, which were read verbatim according to standardized guidelines. Calls were conducted in quiet environments using headsets, and background noise was minimized to support audio quality. This protocol ensured consistency across interviewers while preserving the ecological validity of spontaneous speech. Supervisors reviewed a subset of recordings for quality assurance and adherence to conversational and ethical standards. The SCs were already known to participants through their community roles, minimizing the need for additional rapport-building and ensuring natural, consistent conversational engagement.

The system captures both the audio recordings and the transcribed responses. All staff administering the interviews received interactive training to ensure consistency in conducting semi-structured interviews and administering validated scales. The study population consisted of adults aged 60 and above who were able to converse in English and were current residents or members of Klaatch's partner organizations. These partner organizations are community-based facilities offering progressive living arrangements (independent and assisted living options), with assigned, on-site case workers conducting administrative management and semi-regular wellness check-ins. Building on this dataset, we extracted linguistic features to identify new markers that may improve the inference of loneliness. While we use the CEL score as the reference measure of loneliness, our primary aim is to estimate the likelihood of loneliness from linguistic indicators. This approach serves as a foundational step toward validating and enhancing loneliness assessment models.

The data used for this study includes 1465 interviews collected from 310 unique older adults (between ages 53-103 years) from January 2021 to November 2024. The interviews were semi-structured and conducted within the context of broader wellness check-ins. As part of these sessions, the Campaign to End Loneliness Measurement Tool was administered to assess emotional loneliness. This tool includes three standardized questions: (1) I am content with my friendships and relationships, (2) I have enough people I feel comfortable asking for help at any time, (3) My relationships are as satisfying as I would want them to be. The response to these questions range from "Strongly Agree" (scored 0) to "Strongly Disagree" (scored 4), with scores summed to produce a range from 0 to 12. These responses served as the basis for quantifying each participant's level of emotional loneliness and were used to compute the CEL score [24]. However, the explicit presence of response terms like "agree," "disagree," "strongly agree," or "strongly disagree" in the interview could act as direct cues for the model, potentially giving away the loneliness level based on those words alone, rather than learning meaningful linguistic and acoustic patterns. To prevent this label leakage and make the model learn genuine indicators of loneliness, we removed all such words and their variants from the interview transcripts prior to feature extraction and modeling. We also removed words like 'speaker' and 'Lisa', from the transcripts, which are used in reference to the interviewer.From the audio files available, we masked the portions where explicit survey response words (e.g., "agree," "disagree," "strongly agree," "strongly disagree") were spoken by replacing them with zero-amplitude samples, while preserving the original timing and alignment of the audio signal. This masking approach was chosen to prevent direct label leakage while minimizing distortion of temporal and prosodic structure (e.g., speech rate and pause patterns).

Feature extraction was performed using windowed analyses (25-millisecond window) and internal normalization and aggregation (e.g., frame-level statistics summarized across the full recording), limiting the influence of silent frames on voiced-speech descriptors. Zero-amplitude regions therefore contribute minimally to spectral and prosodic estimates such as pitch, formants, or spectral contrast, rather than introducing spurious values. Table 1 displays participant information.

***Sampling via propensity score matching***

In our analysis of loneliness, we adopted a two-tiered approach to account for both population-wide trends and subgroup-specific variations. For the primary analysis, aimed at understanding general patterns of loneliness, we utilized the complete dataset comprising all available participants. This comprehensive use of the data allowed us to maximize statistical power and capture the full range of variability in loneliness-related responses across the population.

However, when focusing on specific subgroups (e.g., based on gender and race), direct comparisons using the full data could lead to biased or misleading conclusions due to imbalances in sample size and confounding covariates. To address this, we applied propensity score matching [25], using logistic regression, before conducting subgroup-level analyses. Specifically, we treated the male participants as the reference group and matched them to a comparable control group of an equal number of females using age, race, and total CEL score as covariates. Similarly, we found an equal number of subgroup data for race based on age, gender, and total CEL score. This suggests that the subgroup analyses were not confounded by differences in sample characteristics.

The matched samples allowed us to generate more accurate predictions of loneliness within each subgroup. By controlling for potential selection bias and improving covariate balance, our use of propensity score matching enhanced the validity of our inferences regarding subgroup-specific loneliness patterns.

***Identifying features for loneliness (higher end of the CEL score scale)***

*Audio features*

To analyze the audio features associated with loneliness, we extracted audio features from the interview recordings available for all 1,465 interviews. Feature extraction was performed using a combination of established Python toolkits like OpenSMILE, Librosa and Whisper along with models to capture a diverse range of acoustic properties. We computed two types of embeddings from Whisper: whisper-mean and whisper-median. These embeddings provided a compact yet informative summary of the speech content and delivery style.

*Text features*

In addition to audio features, we conducted a comprehensive linguistic analysis of the text data associated with all 1,465 entries in the dataset. For this purpose, we utilized the Differential Language Analysis Toolkit (DLATK) [26], a robust and scalable framework for large-scale language processing and feature extraction.

We extracted the textual features in the following categories:

Dictionary-based approach:

Linguistic Inquiry and Word Count (LIWC) 2022 [27]: This lexicon-based tool captures psychological, emotional, and syntactic dimensions of language by mapping words to predefined psychological categories (e.g., affect, cognition, emotion, tone).

Open-vocabulary approach:

We used Latent Dirichlet Allocation (LDA) to generate 100 topics, capturing thematic patterns in language use across the corpus. This number was chosen to balance interpretability and topic coherence while avoiding excessive fragmentation of topics [28]. LDA's probabilistic framework provides clearer topic separation compared to non-probabilistic methods like keyword clustering, offering transparent word-to-topic and topic-to-document distributions that support interpretable and meaningful insights.

N-grams

We included unigrams, bigrams, and trigrams to represent the frequency and structure of common word patterns and phrases in the text.

All textual data underwent preprocessing in accordance with DLATK's default pipeline, including tokenization and lowercasing. For each feature set (LIWC, LDA topics, and N-grams), we conducted correlation analysis, calculating Pearson correlation coefficients between each feature and the CEL score, to examine their individual associations with emotional loneliness. Bonferroni adjustment was used to correct for multiple comparisons. This text analysis allowed us to capture both psychologically interpretable and statistically robust patterns in language that may reflect or predict loneliness levels.

For the audio analysis, all the audio data also underwent certain preprocessing steps. The audio sections of interviewer asking questions was silenced. Along with that, sections of the audio that included words like "agree", "disagree", "Strongly agree", "Strongly disagree" among others were also muted.

Finally, to predict loneliness based on these extracted audio and text features, we used the ExtraTrees regression model [29]. ExtraTrees was chosen as it reduces variance and captures complex, nonlinear relationships between predictors and the outcome. We used the ExtraTrees Regressor, with 5-fold cross-validation to mitigate overfitting. The cross-validation splits were performed at the participant level, ensuring that interviews from the same individual did not appear in both training and test folds. Model performance was reported via Pearson's r scores.

The primary goal of this analysis was to identify which vocal and linguistic features are most predictive of loneliness. For each modality (audio, text, and combined), we trained the ExtraTrees regressor using the full set of features, then evaluated the model's performance using Pearson correlation between predicted and actual CEL scores. To identify which feature had the greatest influence on predicting loneliness, we examined the feature importance scores generated by the ExtraTrees regression model. These scores reflect how each feature contributed to decision-making across all trees in the ensemble, allowing us to determine which linguistic and acoustic signals were most strongly associated with emotional loneliness.

All analyses were conducted in Python (v3.10) using scikit-learn, DLATK, Librosa, and OpenSMILE libraries. Code for feature extraction, modeling, and validation will be released in the project's GitHub repository.

## Data Availability

The raw data that support the findings of this study is available with Klaatch.

**Code Availability**

The code used for data preprocessing, feature extraction, and statistical modeling in this study is publicly available at:

https://github.com/karthik-strikes/Audio_Analysis

The repository includes documentation and scripts sufficient to reproduce the analyses reported in this manuscript.

**Acknowledgments:**

Dr. Cudjoe was supported by the Johns Hopkins University Center for Innovative Medicine Human Aging Project as a Caryl & George Bernstein Scholar, and the Robert and Jane Meyerhoff Endowed Professorship.

**Author contributions:**

Conceptualization, Supervision: A.G., S.C.G., Data-collection: A.G., E.D., E.A., Formal Analysis: S.K.K., V.K., Writing-original draft: V.K., Review and Editing: N.S., Review: M.L.S., T.K.M.C.

**Competing Interests**

Dr. Cudjoe reported receiving personal fees from Edenbridge Healthcare and Papa, Inc. outside the submitted work. The others authors declare no competing financial or non-financial interests.

**Table 1: Participant Demographics**

| Category | Group | Number of Interviews (Unique Participants) | Recordings /Participant (Mean) | CEL Total (0-12) (Mean) | Mean Age (SD) | Mean Words per Interview (SD) |
|---|---|---|---|---|---|---|
| Overall | All | 1465 (310) | 4.73 | 2.57 | 80.89 (10.15) | 1308.77 (1128.82) |
| Gender | Female | 934 (168) | 5.56 | 2.12 | 81.83 (9.87) | 1304.38 (1120.6) |
| | Male | 406 (91) | 4.46 | 2.58 | 77.81 (8.68) | 1428.86 (1213.34) |
| | Unknown | 125 (51) | - | - | - | 951.51 (781.93) |
| Race | White | 924 (181) | 5.10 | 2.73 | 81.17 (9.96) | 1416.95 (1210.63) |
| | Black | 383 (70) | 5.47 | 2.22 | 79.25 (9.08) | 1135.63 (933.51) |
| | Unknown | 158 (59) | - | - | 87.74 (15.88) | 1095.81 (978.62) |
| Education | Bachelor's Degree or Higher | 358 (75) | 4.77 | 2.66 | 81.41 (8.23) | 1615.64 (1422.58) |
| | Some College or Associate's Degree | 155 (25) | 6.2 | 2.55 | 76.92 (8.11) | 1548.58 (1056.18) |
| | High School or Below | 397 (63) | 6.3 | 2.24 | 80.77 (10.26) | 1172.31 (919.03) |
| | Unknown | 555 (147) | 3.78 | 2.76 | 81.95 (11.65) | 1141.46 (1015.38) |
| Monthly Income | Low income ($0 - $1000) | 271 (60) | 4.52 | 2.54 | 79.32 (11.47) | 1642.94 (1438.77) |
| | Mid-level income ($1000 - $3000) | 946 (167) | 5.66 | 2.48 | 80.79 (9.56) | 1279.57 (1064.84) |
| | High income ($3000 - $6500) | 144 (36) | 4 | 2.68 | 84.49 (10.43) | 1114.17 (923.24) |
| | Unknown | 104 (47) | 2.21 | 3.35 | - | 973.06 (793.78) |
| Living with | Self | 1244 (233) | 5.34 | 3.23 | 84.70 (15.38) | 1091.32 (968.09) |
| | Domestic care provider | 46 (14) | 3.29 | 2.78 | 76.15 (8.64) | 818 (505.49) |
| | Unknown | 175 (63) | 2.77 | 3.23 | 84.70 (15.38) | 1091.32 (968.09) |
| Partner status | Single | 477 (79) | 6.04 | 2.47 | 76.67 (7.81) | 1563.52 (1308.59) |
| | Widow/Divorced | 745 (146) | 5.10 | 2.43 | 83.73 (9.89) | 1229.01 (1043.57) |
| | Married/Domestic Partner | 43 (13) | 3.31 | 2.72 | 77.51 (8.91) | 984.56 (527.09) |
| | Unknown | 200 (72) | 2.78 | 3.32 | 81.36 (14.49) | 1068 (942.19) |

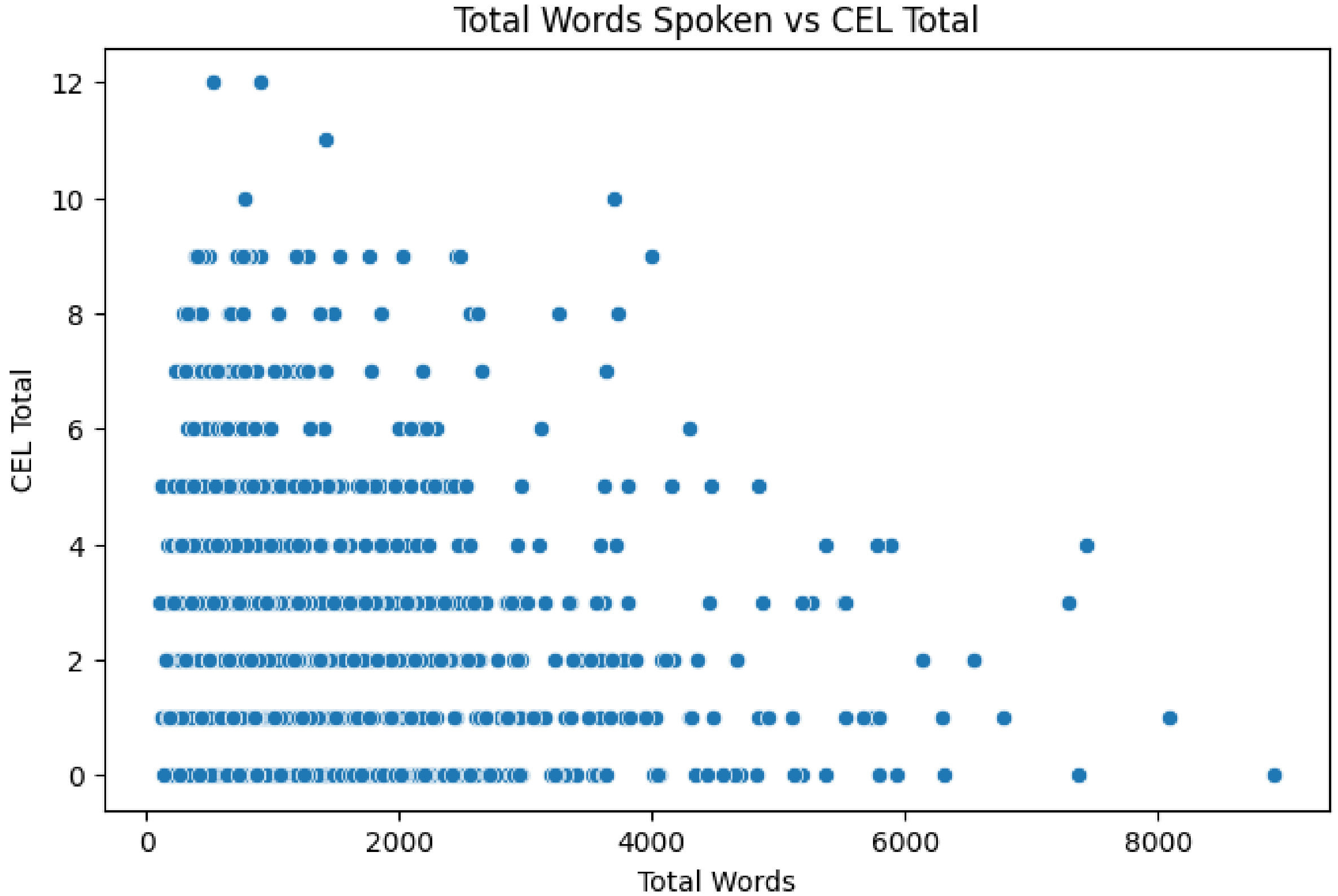

Fig 1: Total Words Spoken by interviewee vs CEL Total Score

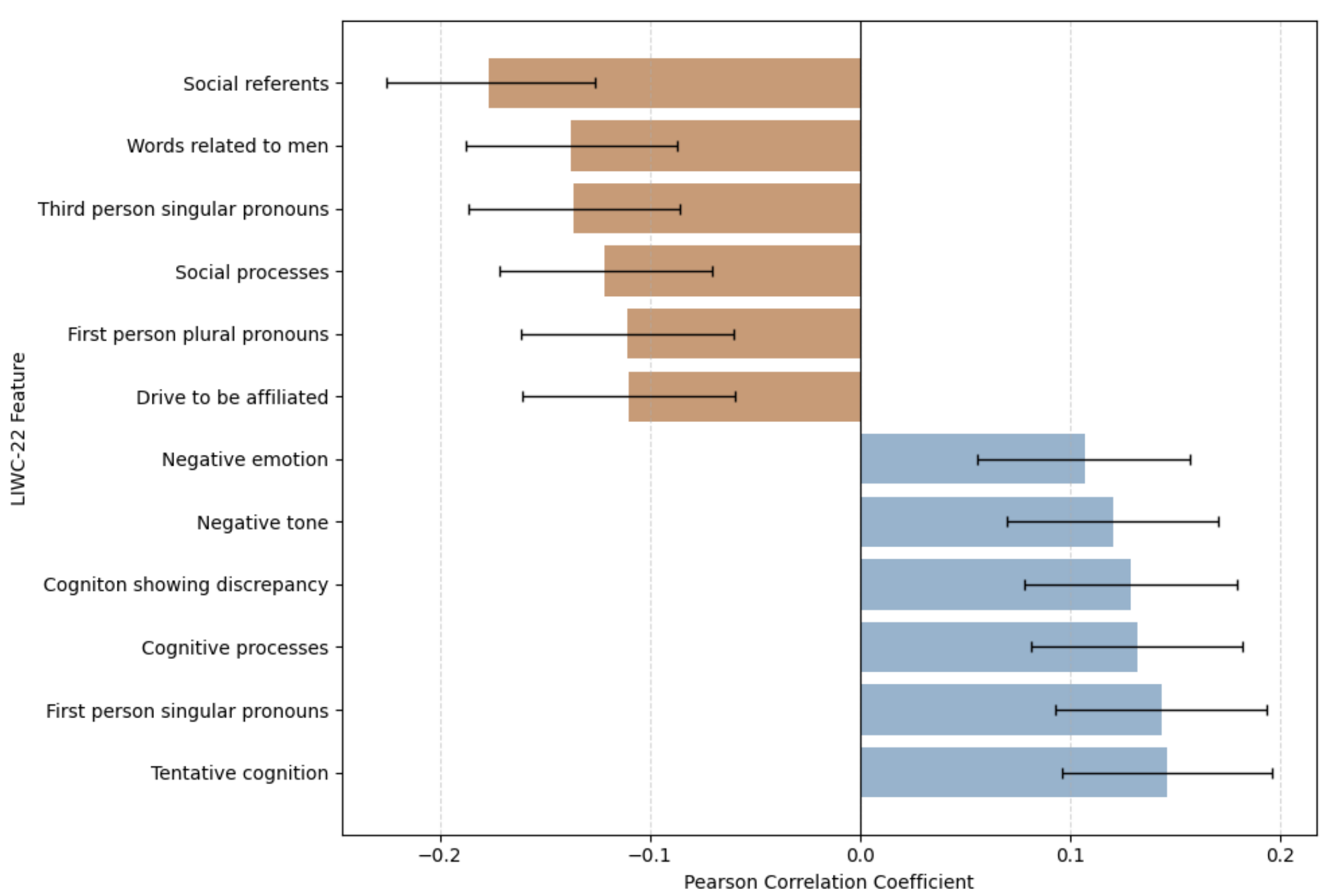

Fig 2: LIWC Correlation with Loneliness (Complete Corpus) (Error bars indicate 95% confidence intervals. All comparisons p<.001.)

**Table 2: LDA Topic Correlations with Loneliness (Complete Corpus)**

Negative correlations indicate a higher correlation with low loneliness, positive indicates higher correlation with high loneliness.

| Top 10 words in topic | Pearson r | p-value | N | 95% CI (lower) | 95% CI (upper) |
|---|---|---|---|---|---|
| he's, him, his, six, father, he'll, lives, gone, wife, wants | -0.11 | 0.00 | 1465 | -0.16 | -0.06 |
| said, told, came, put, doctor, says, door, asked, saw, floor | -0.11 | 0.00 | 1465 | -0.16 | -0.06 |
| kids, his, son, daughter, school, love, house, dad, mother, married | -0.10 | 0.00 | 1465 | -0.15 | -0.05 |
| listen, grandma, grandkids, granddaughter, relationship, mom, speak, ain't, miss, yourself | -0.10 | 0.00 | 1465 | -0.15 | -0.04 |
| richard, exactly, drink, calls, stephen, bloody, wish, type, bah, anyhow | -0.09 | 0.00 | 1465 | -0.14 | -0.04 |
| kids, basically, hours, yale, eight, providence, full, huge, laundry, parents | -0.09 | 0.00 | 1465 | -0.14 | -0.04 |
| eat, food, put, make, cook, chicken, salad, sauce, dinner, cooking | -0.09 | 0.00 | 1465 | -0.14 | -0.04 |
| work, done, working, construction, security, rest, retired, it'll, they'll, sleep | -0.09 | 0.01 | 1465 | -0.14 | -0.03 |
| come, let's, tv, watch, hold, sorry, question, tomorrow, number, answer | 0.10 | 0.00 | 1465 | 0.05 | 0.15 |
| little, talk, sometimes, bit, myself, sister, music, exercise, along, sometime | 0.11 | 0.00 | 1465 | 0.06 | 0.16 |
| sister, nursing, nurse, deal, difficult, candy, although, frequently, security, writing | 0.12 | 0.00 | 1465 | 0.07 | 0.17 |
| think, has, close, bit, quite, seem, seems, felt, feeling, they've | 0.12 | 0.00 | 1465 | 0.07 | 0.17 |
| told, pay, bill, utopia, anybody, seen, understand, knows, hell, security | 0.12 | 0.00 | 1465 | 0.07 | 0.17 |
| also, group, being, find, speak, jewish, basically, residents, understand, activities | 0.14 | 0.00 | 1465 | 0.08 | 0.1 |

**Table 3: Audio Feature Correlations with Loneliness (Complete Corpus)**
Negative correlations indicate a higher correlation with low loneliness, positive indicates higher correlation with high loneliness.

| Features | Pearson r | p-value | N | 95% CI (lower) | 95% CI (upper) |
|---|---|---|---|---|---|
| Librosa | | | | | |
| Contrast in the second frequency sub-band | -0.23 | 0.00 | 1465 | -0.28 | -0.18 |
| Contrast in the third frequency sub-band | -0.20 | 0.00 | 1465 | -0.25 | -0.15 |
| Overall spectral slope of the audio signal | -0.17 | 0.00 | 1465 | -0.22 | -0.12 |
| Contrast in the fourth frequency sub-band | -0.14 | 0.00 | 1465 | -0.19 | -0.09 |
| First formant, related to vowel sounds in speech | -0.13 | 0.00 | 1465 | -0.18 | -0.08 |
| Higher-frequency spectral nuances | 0.15 | 0.00 | 1465 | 0.10 | 0.20 |
| Energy in pitch class B | 0.16 | 0.00 | 1465 | 0.11 | 0.21 |
| Intricate spectral patterns, useful for speaker identification | 0.17 | 0.00 | 1465 | 0.12 | 0.22 |
| Opensmile | | | | | |
| Mean log amplitude of the 3rd formant (F3) relative to fundamental frequency | -0.24 | 0.00 | 1465 | -0.28 | -0.19 |
| Number of loudness peaks per second, indicating dynamic changes | -0.23 | 0.00 | 1465 | -0.28 | -0.18 |
| Mean log amplitude of the 1st formant (F1) relative to fundamental frequency | -0.23 | 0.00 | 1465 | -0.28 | -0.18 |
| Mean log amplitude of the 2nd formant (F2) relative to fundamental frequency | -0.23 | 0.00 | 1465 | -0.28 | -0.19 |
| Median loudness distribution | -0.19 | 0.00 | 1465 | -0.24 | -0.14 |
| Normalized standard deviation of loudness | 0.16 | 0.00 | 1465 | 0.11 | 0.21 |
| Normalized standard deviation of overall spectral flux | 0.18 | 0.00 | 1465 | 0.13 | 0.23 |
| Normalized standard deviation of F1 amplitude relative to fundamental frequency | 0.22 | 0.00 | 1465 | 0.18 | 0.27 |
| Normalized standard deviation of F3 amplitude relative to fundamental frequency | 0.23 | 0.00 | 1465 | 0.18 | 0.28 |
| Normalized standard deviation of F2 amplitude relative to fundamental frequency | 0.23 | 0.00 | 1465 | 0.18 | 0.27 |

**Table 4: Predictions (Complete Corpus) using Extra Trees showing Pearson correlations with CEL score.**

| **Features** | **Complete Corpus** | **Male** | **Female** | **White** | **Black** |
|---|---:|---:|---:|---:|---:|
| Librosa | 0.179 | 0.037 | 0.105 | 0.181 | 0.160 |
| Opensmile | 0.173 | 0.131 | 0.121 | 0.299 | 0.238 |
| Whisper (mean model) | 0.125 | 0.07 | 0.096 | 0.140 | 0.069 |
| Whisper (median model) | 0.108 | 0.019 | 0.127 | 0.095 | 0.053 |
| Combined (Audio) | 0.195 | 0.141 | 0.224 | 0.286 | 0.190 |
| | | | | | |
| LIWC | 0.269 | 0.066 | 0.207 | 0.245 | 0.175 |
| LDA 100 TOPICS | 0.183 | 0.118 | 0.174 | 0.188 | 0.083 |
| n-grams | 0.258 | 0.274 | 0.199 | 0.097 | 0.167 |
| Combined (Text) | 0.283 | 0.234 | 0.229 | 0.17 | 0.166 |
| | | | | | |
| Overall Combined | 0.298 | 0.230 | 0.228 | 0.343 | 0.219 |

**Supplementary Information:**

**Supplementary Information A: Interviewer Training and Data Collection Procedures:**
All interviews were conducted by onsite SCs employed within HUD Section 202 senior housing communities. SCs met federal HUD-aligned qualifications, including a bachelor's degree (or equivalent experience) in social work, psychology, gerontology, or a related human services field, and prior experience working with older adults. Before data collection, SCs completed standardized onboarding and refresher training focused on ethical data collection, conversational interviewing techniques, and consistent administration of survey instruments. Supervisory staff periodically reviewed recordings to ensure adherence to protocol and data quality.

At the start of each call, participants provided verbal consent for audio recording and participation. Participation was voluntary, and individuals could decline to answer any question or withdraw at any time without penalty. No financial incentives were offered. Interviews were conducted in English and followed a standardized structure: (1) an open-ended conversational segment designed to elicit natural speech, (2) administration of the Campaign to End Loneliness (CEL) items, (3) engagement-related questions, and (4) an optional staff assessment.

To promote naturalistic language and reduce stigma, interviewers were instructed not to use the term "loneliness" during the conversational portion of the interview. Instead, loneliness was assessed exclusively through the validated CEL items administered after the conversation. Open-ended prompts (e.g., "Tell me about your day," "Can you share a story about something you enjoy?") were followed by neutral follow-up questions to encourage elaboration without steering content.

SCs were trained to minimize background noise, conduct calls from quiet environments, and ensure adequate recording quality. All interviews were recorded using a secure telephone platform. Audio and survey data were de-identified prior to analysis and stored on encrypted servers with access restricted to authorized personnel.

**Table S1: LIWC Complete Corpus**
Negative correlations indicate a higher correlation with low loneliness, positive indicates higher correlation with high loneliness ($p \le 0.05$).

| **Category** | **Pearson r** | **p-value** | **N** | **95% CI (lower)** | **95% CI (upper)** |
|---|---|---|---|---|---|
| Social referents | -0.18 | 0.00 | 1465 | -0.23 | -0.13 |
| Words related to men | -0.14 | 0.00 | 1465 | -0.19 | -0.09 |
| Third person singular pronouns | -0.14 | 0.00 | 1465 | -0.19 | -0.09 |
| Social processes | -0.12 | 0.00 | 1465 | -0.17 | -0.07 |
| First person plural pronouns | -0.11 | 0.00 | 1465 | -0.16 | -0.06 |
| Drive to be affiliated | -0.11 | 0.00 | 1465 | -0.16 | -0.06 |
| Conjunctions | -0.11 | 0.00 | 1465 | -0.16 | -0.05 |
| Drives | -0.10 | 0.00 | 1465 | -0.15 | -0.05 |
| Perception of motion | -0.09 | 0.00 | 1465 | -0.14 | -0.04 |
| Words related to women | -0.09 | 0.00 | 1465 | -0.14 | -0.04 |
| Words referring to family | -0.09 | 0.00 | 1465 | -0.14 | -0.04 |

| Words related to food | -0.08 | 0.01 | 1465 | -0.13 | -0.03 |
|---|---|---|---|---|---|
| Second person pronouns | -0.08 | 0.01 | 1465 | -0.13 | -0.03 |
| Religion | -0.07 | 0.02 | 1465 | -0.12 | -0.02 |
| Cognition representing all or none approach | -0.07 | 0.02 | 1465 | -0.12 | -0.02 |
| Physical | -0.07 | 0.02 | 1465 | -0.12 | -0.02 |
| Acquire | -0.07 | 0.03 | 1465 | -0.12 | -0.02 |
| Perception of Time | -0.07 | 0.03 | 1465 | -0.12 | -0.02 |
| Prosocial behavior | 0.07 | 0.03 | 1465 | 0.02 | 0.12 |
| Perception of feeling | 0.07 | 0.02 | 1465 | 0.02 | 0.12 |
| Social behavior | 0.07 | 0.02 | 1465 | 0.02 | 0.12 |
| Cognition | 0.08 | 0.01 | 1465 | 0.03 | 0.13 |
| Differentiation | 0.08 | 0.01 | 1465 | 0.03 | 0.13 |
| Memory | 0.08 | 0.01 | 1465 | 0.03 | 0.13 |
| Communication | 0.08 | 0.01 | 1465 | 0.03 | 0.13 |
| Polite social behavior | 0.08 | 0.01 | 1465 | 0.03 | 0.13 |
| Common adjectives | 0.10 | 0.00 | 1465 | 0.05 | 0.15 |
| Negative emotion | 0.11 | 0.00 | 1465 | 0.06 | 0.16 |
| Negative tone | 0.12 | 0.00 | 1465 | 0.07 | 0.17 |
| Cognition showing discrepancy | 0.13 | 0.00 | 1465 | 0.08 | 0.18 |
| Cognitive processes | 0.13 | 0.00 | 1465 | 0.08 | 0.18 |
| First person singular pronouns | 0.14 | 0.00 | 1465 | 0.09 | 0.19 |
| Tentative cognition | 0.15 | 0.00 | 1465 | 0.10 | 0.20 |

**Table S2: LIWC Correlations with Loneliness by Gender**

Negative correlations indicate a higher correlation with low loneliness, positive indicates higher correlation with high loneliness ($p \leq 0.05$).

| Category | Pearson r | p-value | N | 95% CI (lower) | 95% CI (upper) |
|---|---|---|---|---|---|
| Male | | | | | |
| Third person singular pronouns | -0.17 | 0.02 | 397 | -0.26 | -0.07 |
| Social referents | -0.16 | 0.03 | 397 | -0.25 | -0.06 |
| Focus in the future | -0.15 | 0.05 | 397 | -0.24 | -0.05 |
| Second person pronouns | -0.14 | 0.05 | 397 | -0.24 | -0.05 |
| Social processes | -0.14 | 0.05 | 397 | -0.24 | -0.04 |
| Common adjectives | 0.14 | 0.05 | 397 | 0.05 | 0.24 |
| Cognitive processes | 0.15 | 0.04 | 397 | 0.05 | 0.25 |
| Cognition | 0.16 | 0.03 | 397 | 0.06 | 0.26 |
| Cognition showing discrepancy | 0.21 | 0 | 397 | 0.11 | 0.3 |
| Tentative cognition | 0.23 | 0 | 397 | 0.13 | 0.32 |
| Female | | | | | |
| Third person singular pronouns | -0.21 | 0 | 397 | -0.31 | -0.12 |
| Social referents | -0.21 | 0 | 397 | -0.31 | -0.12 |
| Words related to men | -0.19 | 0 | 397 | -0.29 | -0.1 |
| Focus in the past | -0.19 | 0 | 397 | -0.28 | -0.09 |
| Conjunctions | -0.17 | 0.01 | 397 | -0.27 | -0.07 |
| Social processes | -0.17 | 0.01 | 397 | -0.26 | -0.07 |
| Religion | -0.17 | 0.01 | 397 | -0.26 | -0.07 |
| Memory | 0.17 | 0.01 | 397 | 0.08 | 0.27 |
| Cognitive processes | 0.17 | 0.01 | 397 | 0.08 | 0.27 |
| Common adjectives | 0.18 | 0.01 | 397 | 0.08 | 0.27 |
| Tentative cognition | 0.2 | 0 | 397 | 0.1 | 0.29 |

| First person singular pronouns | 0.21 | 0 | 397 | 0.12 | 0.3 |
|---|---|---|---|---|---|

**Table S3: LIWC Correlations with Loneliness by Race**

Negative correlations indicate a higher correlation with low loneliness, positive indicates higher correlation with high loneliness ($p \leq 0.05$).

| **Category** | **Pearson r** | **p-value** | **N** | **95% CI (lower)** | **95% CI (upper)** |
|---|---|---|---|---|---|
| White | | | | | |
| Social referents | -0.18 | 0.01 | 383 | -0.27 | -0.08 |
| Cognitive processes | 0.16 | 0.02 | 383 | 0.06 | 0.26 |
| Cognition showing discrepancy | 0.17 | 0.01 | 383 | 0.07 | 0.27 |
| Tentative cognition | 0.17 | 0.01 | 383 | 0.07 | 0.27 |
| Perception of feeling | 0.18 | 0.01 | 383 | 0.08 | 0.27 |
| First person singular pronouns | 0.18 | 0.01 | 383 | 0.08 | 0.28 |
| Negative emotion | 0.19 | 0.01 | 383 | 0.09 | 0.28 |
| Negative tone | 0.24 | 0.00 | 383 | 0.14 | 0.33 |
| Black | | | | | |
| Focus in the past | -0.20 | 0.01 | 383 | -0.29 | -0.10 |
| Third person singular pronouns | -0.19 | 0.01 | 383 | -0.29 | -0.09 |

**Table S4: LDA Topic Correlations with Loneliness (Complete Corpus)**
Negative correlations indicate a higher correlation with low loneliness, positive indicates higher correlation with high loneliness ($p \leq 0.05$).

| Top 10 words in topic | Pearson r | p-value | N | 95% CI (lower) | 95% CI (upper) |
|---|---|---|---|---|---|
| he's, him, his, six, father, he'll, lives, gone, wife, wants | -0.11 | 0.00 | 1465 | -0.16 | -0.06 |
| said, told, came, put, doctor, says, door, asked, saw, floor | -0.11 | 0.00 | 1465 | -0.16 | -0.06 |
| kids, his, son, daughter, school, love, house, dad, mother, married | -0.10 | 0.00 | 1465 | -0.15 | -0.05 |
| listen, grandma, grandkids, granddaughter, relationship, mom, speak, ain't, miss, yourself | -0.10 | 0.00 | 1465 | -0.15 | -0.04 |
| richard, exactly, drink, calls, stephen, bloody, wish, type, bah, anyhow | -0.09 | 0.00 | 1465 | -0.14 | -0.04 |
| kids, basically, hours, yale, eight, providence, full, huge, laundry, parents | -0.09 | 0.00 | 1465 | -0.14 | -0.04 |
| eat, food, put, make, cook, chicken, salad, sauce, dinner, cooking | -0.09 | 0.00 | 1465 | -0.14 | -0.04 |
| work, done, working, construction, security, rest, retired, it'll, they'll, sleep | -0.09 | 0.01 | 1465 | -0.14 | -0.03 |
| mm, hmm, huh, uh-huh, mm-hmm, wow, connecticut, winter, face, thanksgiving | -0.08 | 0.01 | 1465 | -0.13 | -0.03 |
| her, she's, has, told, daughter, doesn't, name, goes, coming, comes | -0.08 | 0.01 | 1465 | -0.13 | -0.03 |
| god, heart, john, elizabeth, failure, saint, yea, dear, hey, she'd | -0.08 | 0.01 | 1465 | -0.13 | -0.03 |
| every, day, morning, once, lunch, downstairs, walk, dinner, usually, 00 | -0.08 | 0.01 | 1465 | -0.13 | -0.03 |
| wonderful, father, children, parents, soup, garden, afternoon, st, television, turn | -0.08 | 0.01 | 1465 | -0.13 | -0.02 |
| sister, house, brother, mother, sisters, bring, girls, ready, christmas, older | -0.07 | 0.02 | 1465 | -0.12 | -0.02 |
| church, sunday, bible, service, saturday, friday, pastor, study, monday, lord | -0.07 | 0.03 | 1465 | -0.12 | -0.02 |
| mean, i've, haven't, actually, done, point, best, able, works, makes | 0.07 | 0.04 | 1465 | 0.01 | 0.12 |
| problem, goes, deal, social, number, we'll, da, everyone, reach, move | 0.07 | 0.02 | 1465 | 0.02 | 0.12 |
| times, probably, group, usually, relationships, exactly, message, content, please, program | 0.07 | 0.02 | 1465 | 0.02 | 0.12 |
| school, bike, yale, vote, chicago, open, war, speak, state, speaking | 0.08 | 0.01 | 1465 | 0.03 | 0.13 |
| stuff, huh, va, problems, paid, deal, business, system, they'll, somewhere | 0.08 | 0.01 | 1465 | 0.03 | 0.13 |
| week, times, three, days, four, weeks, usually, least, ten, contact | 0.08 | 0.01 | 1465 | 0.03 | 0.13 |
| if, something, way, could, feel, anything, ask, tell, let, saying | 0.08 | 0.01 | 1465 | 0.03 | 0.13 |
| ok, relationships, relationship, o'clock, hold, pandemic, thirty, twenty, wedding, patricia | 0.09 | 0.00 | 1465 | 0.04 | 0.14 |
| train, age, we've, coffee, appreciate, set, trump, cafe, isn't, mary | 0.09 | 0.00 | 1465 | 0.04 | 0.14 |
| thank, friends, hello, sorry, ahead, hi, relationship, ten, twice, mostly | 0.10 | 0.00 | 1465 | 0.05 | 0.15 |
| come, let's, tv, watch, hold, sorry, question, tomorrow, number, answer | 0.10 | 0.00 | 1465 | 0.05 | 0.15 |
| little, talk, sometimes, bit, myself, sister, music, exercise, along, sometime | 0.11 | 0.00 | 1465 | 0.06 | 0.16 |
| sister, nursing, nurse, deal, difficult, candy, although, frequently, security, writing | 0.12 | 0.00 | 1465 | 0.07 | 0.17 |
| think, has, close, bit, quite, seem, seems, felt, feeling, they've | 0.12 | 0.00 | 1465 | 0.07 | 0.17 |
| told, pay, bill, utopia, anybody, seen, understand, knows, hell, security | 0.12 | 0.00 | 1465 | 0.07 | 0.17 |
| also, group, being, find, speak, jewish, basically, residents, understand, activities | 0.14 | 0.00 | 1465 | 0.08 | 0.19 |

**Table S5: LDA Topic Correlations with Loneliness by Gender**

Negative correlations indicate a higher correlation with low loneliness, positive indicates higher correlation with high loneliness ($p \leq 0.05$).

| **Top 10 words in topic** | **Pearson r** | **p-value** | **N** | **95% CI (lower)** | **95% CI (upper)** |
|---|---|---|---|---|---|
| Male | | | | | |
| this, said, going, be, mean, if, want, would, say, i'll | -0.16 | 0.03 | 397 | -0.26 | -0.07 |
| mean, anyway, big, york, wife, dinner, limited, enjoy, trivia, jersey | -0.15 | 0.04 | 397 | -0.24 | -0.05 |
| day, every, daughter, call, two, by, times, three, sit, walk | -0.14 | 0.05 | 397 | -0.23 | -0.04 |
| very, great, life, wonderful, play, love, couldn't, young, wife, heart | -0.14 | 0.05 | 397 | -0.23 | -0.04 |
| swimming, pool, haven, close, florida, bike, open, georgia, trip, exercise | 0.14 | 0.05 | 397 | 0.04 | 0.24 |
| as, much, kind, things, different, used, than, many, pretty, working | 0.15 | 0.04 | 397 | 0.05 | 0.24 |
| okay, huh, hello, thank, mhm, bye, watch, leave, might, lady | 0.15 | 0.04 | 397 | 0.05 | 0.24 |
| they'll, certain, comes, paid, vision, none, machines, kinds, cause, occasionally | 0.16 | 0.03 | 397 | 0.06 | 0.25 |
| had, would, used, were, didn't, over, family, many, place, took | 0.16 | 0.03 | 397 | 0.06 | 0.25 |
| stuff, put, huh, most, problems, enough, va, goes, us, set | 0.17 | 0.02 | 397 | 0.07 | 0.26 |
| myself, enough, doctor, house, bank, man, lose, pain, card, taken | 0.19 | 0.01 | 397 | 0.09 | 0.28 |
| sister, nurse, someone, comes, weeks, nursing, awful, although, takes, nurses | 0.23 | 0.00 | 397 | 0.13 | 0.32 |
| Female | | | | | |
| going, her, got, up, get, there, this, if, at, then | -0.18 | 0.01 | 397 | -0.28 | -0.09 |
| church, long, sunday, stuff, yep, pastor, supposed, needs, study, richardson | -0.18 | 0.01 | 397 | -0.27 | -0.08 |
| went, down, nice, around, used, worked, ago, doctor, each, working | -0.18 | 0.01 | 397 | -0.27 | -0.08 |
| richard, yep, exactly, . ., bloody, calls, stephen, anyhow, type, jeep | -0.17 | 0.01 | 397 | -0.26 | -0.07 |
| then, listen, other, than, relationship, need, grandma, let, sisters, through | -0.15 | 0.03 | 397 | -0.25 | -0.05 |
| say, your, family, tell, things, life, their, love, everybody, phone | -0.15 | 0.03 | 397 | -0.24 | -0.05 |
| home, stuff, mother, walk, may, us, even, person, our, nobody | -0.14 | 0.03 | 397 | -0.24 | -0.05 |

| book, actually, theater, show, extremely, club, pick, sarah, synagogue, bunch | 0.15 | 0.03 | 397 | 0.05 | 0.24 |
|---|---|---|---|---|---|
| help, feel, family, bye, enjoy, comfortable, ask, work, relationship, again | 0.15 | 0.03 | 397 | 0.05 | 0.24 |
| really, feel, thought, pretty, sometimes, maybe, getting, sort, question, talking | 0.17 | 0.01 | 397 | 0.07 | 0.26 |
| yale, pay, towers, since, understand, find, hold, saturday, bus, please | 0.17 | 0.01 | 397 | 0.07 | 0.26 |
| karen, remember, once, often, donna, offhand, far, mind, door, anybody | 0.21 | 0.00 | 397 | 0.12 | 0.31 |
| think, very, sure, bye, thank, nice, fine, an, try, kind | 0.21 | 0.00 | 397 | 0.12 | 0.31 |

**Table S6: LDA Topic Correlations with Loneliness by Race**

Negative correlations indicate a higher correlation with low loneliness, positive indicates higher correlation with high loneliness ($p \leq 0.05$).

| Top 10 words in topic | Pearson r | p-value | N | 95% CI (lower) | 95% CI (upper) |
|---|---|---|---|---|---|
| White | | | | | |
| mean, thought, anyway, i'll, friend, maybe, somebody, says, old, i'd | -0.19 | 0.00 | 383 | -0.29 | -0.09 |
| mean, place, wife, anyway, big, best, dinner, limited, enjoy, center | -0.18 | 0.00 | 383 | -0.28 | -0.09 |
| school, kids, these, daughter, state, war, wife, experience, wonderful, high | -0.15 | 0.03 | 383 | -0.25 | -0.05 |
| myself, doctor, gas, card, rather, father, regular, foot, dead, sleep | 0.16 | 0.02 | 383 | 0.06 | 0.26 |
| if, want, didn't, call, feel, any, why, ask, person, next | 0.17 | 0.01 | 383 | 0.07 | 0.27 |
| sister, actually, someone, nurse, nursing, awful, weeks, therapist, able, although | 0.19 | 0.00 | 383 | 0.09 | 0.29 |
| class, therapy, find, felt, pay, physical, older, paint, check, start | 0.19 | 0.00 | 383 | 0.10 | 0.29 |
| blah, towers, bike, open, swimming, gym, black, florida, ymca, yale | 0.22 | 0.00 | 383 | 0.13 | 0.32 |
| feel, help, guess, phone, god, someone, middle, haven't, seems, sisters | 0.25 | 0.00 | 383 | 0.16 | 0.34 |
| Black | | | | | |
| her, she's, take, little, doing, where, daughter, granddaughter, house, off | -0.19 | 0.00 | 383 | -0.29 | -0.10 |
| as, country, who, name, anyway, far, africa, town, land, listen | 0.21 | 0.00 | 383 | 0.11 | 0.30 |
| stuff, lot, used, new, many, most, food, old, let's, va | 0.25 | 0.00 | 383 | 0.15 | 0.34 |

**Table S7: Audio Feature Correlations with Loneliness (Complete Corpus)**
Negative correlations indicate a higher correlation with low loneliness, positive indicates higher correlation with high loneliness ($p \leq 0.05$).

| Features | Pearson r | p-value | N | 95% CI (lower) | 95% CI (upper) |
|---|---|---|---|---|---|
| Librosa | | | | | |
| Contrast in the second frequency sub-band | -0.23 | 0.00 | 1465 | -0.28 | -0.18 |
| Contrast in the third frequency sub-band | -0.20 | 0.00 | 1465 | -0.25 | -0.15 |
| Overall spectral slope of the audio signal | -0.17 | 0.00 | 1465 | -0.22 | -0.12 |
| Contrast in the fourth frequency sub-band | -0.14 | 0.00 | 1465 | -0.19 | -0.09 |
| First formant, related to vowel sounds in speech | -0.13 | 0.00 | 1465 | -0.18 | -0.08 |
| Contrast in the fifth frequency sub-band | -0.12 | 0.00 | 1465 | -0.17 | -0.07 |
| Contrast in the sixth frequency sub-band | -0.08 | 0.00 | 1465 | -0.13 | -0.03 |
| Strength of the major third interval | -0.08 | 0.00 | 1465 | -0.13 | -0.03 |
| Contrast in the highest frequency sub-band | 0.06 | 0.02 | 1465 | 0.01 | 0.11 |
| Minute spectral features, aiding in emotion detection | 0.07 | 0.02 | 1465 | 0.01 | 0.12 |
| Subtle difference in speech articulation | 0.08 | 0.00 | 1465 | 0.03 | 0.13 |
| Higher-order spectral variations | 0.09 | 0.00 | 1465 | 0.04 | 0.14 |
| Finer spectral details beyond the primary formants | 0.11 | 0.00 | 1465 | 0.06 | 0.16 |
| Energy in pitch class D-sharp | 0.11 | 0.00 | 1465 | 0.05 | 0.16 |
| Energy in pitch class D | 0.12 | 0.00 | 1465 | 0.07 | 0.17 |
| Energy in pitch class G | 0.12 | 0.00 | 1465 | 0.07 | 0.17 |
| Vocal tract shape | 0.12 | 0.00 | 1465 | 0.07 | 0.17 |
| Energy in pitch class E | 0.12 | 0.00 | 1465 | 0.07 | 0.17 |
| Energy in pitch class F | 0.13 | 0.00 | 1465 | 0.08 | 0.18 |
| Energy in pitch class C | 0.13 | 0.00 | 1465 | 0.08 | 0.18 |
| Energy in pitch class C-sharp | 0.14 | 0.00 | 1465 | 0.09 | 0.19 |
| Energy in pitch class F-sharp | 0.14 | 0.00 | 1465 | 0.09 | 0.19 |
| Energy in pitch class G-sharp | 0.14 | 0.00 | 1465 | 0.09 | 0.19 |
| Higher-frequency spectral nuances | 0.15 | 0.00 | 1465 | 0.10 | 0.20 |
| Fine spectral details, useful in music analysis | 0.15 | 0.00 | 1465 | 0.10 | 0.20 |
| Energy in pitch class B | 0.16 | 0.00 | 1465 | 0.11 | 0.21 |
| Intricate spectral patterns, useful for speaker identification | 0.17 | 0.00 | 1465 | 0.12 | 0.22 |
| Energy in pitch class A | 0.18 | 0.00 | 1465 | 0.13 | 0.23 |
| Energy in pitch class A-sharp | 0.20 | 0.00 | 1465 | 0.15 | 0.25 |
| Opensmile | | | | | |
| Mean log amplitude of the 3rd formant (F3) relative to fundamental frequency | -0.24 | 0.00 | 1465 | -0.28 | -0.19 |
| Number of loudness peaks per second, indicating dynamic changes | -0.23 | 0.00 | 1465 | -0.28 | -0.18 |
| Mean log amplitude of the 1st formant (F1) relative to fundamental frequency | -0.23 | 0.00 | 1465 | -0.28 | -0.18 |
| Mean log amplitude of the 2nd formant (F2) relative to fundamental frequency | -0.23 | 0.00 | 1465 | -0.28 | -0.19 |
| Median loudness distribution | -0.19 | 0.00 | 1465 | -0.24 | -0.14 |
| Higher loudness levels | -0.18 | 0.00 | 1465 | -0.22 | -0.13 |
| Range between 0th and 2nd percentiles of loudness | -0.18 | 0.00 | 1465 | -0.23 | -0.13 |
| Mean loudness | -0.17 | 0.00 | 1465 | -0.22 | -0.12 |
| Overall mean spectral flux | -0.14 | 0.00 | 1465 | -0.19 | -0.09 |
| Equivalent continuous sound level in dB, representing overall loudness | -0.12 | 0.00 | 1465 | -0.17 | -0.06 |
| Standard deviation of voiced segment lengths in seconds | -0.11 | 0.00 | 1465 | -0.16 | -0.06 |
| Average duration of voiced segments in seconds | -0.10 | 0.00 | 1465 | -0.15 | -0.04 |
| Higher pitch range | -0.10 | 0.00 | 1465 | -0.15 | -0.05 |
| Median of fundamental frequency distribution | -0.09 | 0.00 | 1465 | -0.14 | -0.03 |
| Mean spectral flux in unvoiced regions | -0.09 | 0.00 | 1465 | -0.14 | -0.03 |
| Mean fundamental frequency in semitones above 27.5 Hz | -0.09 | 0.00 | 1465 | -0.14 | -0.04 |
| Mean slope of increasing loudness segments | -0.09 | 0.00 | 1465 | -0.14 | -0.04 |
| Lower pitch range | -0.08 | 0.01 | 1465 | -0.13 | -0.03 |
| Mean slope of decreasing loudness segments | -0.08 | 0.01 | 1465 | -0.13 | -0.03 |
| Mean spectral slope from 0-500 Hz in unvoiced regions | -0.07 | 0.01 | 1465 | -0.13 | -0.02 |
| Mean spectral flux in voiced regions | -0.06 | 0.05 | 1465 | -0.11 | -0.01 |
| Mean of spectral slope across all region | -0.06 | 0.04 | 1465 | -0.11 | -0.01 |
| Mean log difference between fundamental frequency and amplitude of second harmonic | -0.06 | 0.04 | 1465 | -0.12 | -0.01 |
| Mean of finer spectral variation in voiced region | 0.06 | 0.04 | 1465 | 0.01 | 0.12 |
| Normalized standard deviation of F3 frequency | 0.06 | 0.04 | 1465 | 0.01 | 0.11 |
| Normalized standard deviation of spectral slope from 0-500 Hz in voiced regions | 0.07 | 0.02 | 1465 | 0.02 | 0.12 |
| Mean of spectral slope across voiced region | 0.08 | 0.01 | 1465 | 0.02 | 0.13 |
| Mean of finer spectral variation across all region | 0.11 | 0.00 | 1465 | 0.06 | 0.16 |
| Standard deviation of unvoiced segment lengths | 0.13 | 0.00 | 1465 | 0.08 | 0.18 |
| Average duration of unvoiced segments | 0.14 | 0.00 | 1465 | 0.09 | 0.19 |
| Normalized standard deviation of loudness | 0.16 | 0.00 | 1465 | 0.11 | 0.21 |
| Normalized standard deviation of overall spectral flux | 0.18 | 0.00 | 1465 | 0.13 | 0.23 |
| Normalized standard deviation of F1 amplitude relative to fundamental frequency | 0.22 | 0.00 | 1465 | 0.18 | 0.27 |
| Normalized standard deviation of F3 amplitude relative to fundamental frequency | 0.23 | 0.00 | 1465 | 0.18 | 0.28 |
| Normalized standard deviation of F2 amplitude relative to fundamental frequency | 0.23 | 0.00 | 1465 | 0.18 | 0.27 |

**Table S8: Audio Feature Correlations with Loneliness by Gender**

Negative correlations indicate a higher correlation with low loneliness, positive indicates higher correlation with high loneliness ($p \leq 0.05$).

| Features | Pearson r | p-value | N | 95% CI (lower) | 95% CI (upper) |
|---|---|---|---|---|---|
| Librosa (Male) | | | | | |
| Energy in pitch class C | 0.12 | 0.03 | 397 | 0.03 | 0.22 |
| Finer spectral details beyond the primary formants | 0.13 | 0.02 | 397 | 0.03 | 0.23 |
| Vocal tract shape | 0.14 | 0.01 | 397 | 0.05 | 0.24 |
| Energy in pitch class F-sharp | 0.16 | 0.00 | 397 | 0.06 | 0.26 |
| Energy in pitch class G-sharp | 0.16 | 0.01 | 397 | 0.06 | 0.25 |
| Energy in pitch class G | 0.16 | 0.01 | 397 | 0.06 | 0.25 |
| Fine spectral details, useful in music analysis | 0.17 | 0.00 | 397 | 0.07 | 0.26 |
| Energy in pitch class A | 0.20 | 0.00 | 397 | 0.10 | 0.29 |
| Energy in pitch class B | 0.20 | 0.00 | 397 | 0.10 | 0.29 |
| Captures intricate spectral patterns, useful for speaker identification | 0.21 | 0.00 | 397 | 0.11 | 0.30 |
| Captures higher-frequency spectral nuances | 0.22 | 0.00 | 397 | 0.12 | 0.31 |
| Energy in pitch class A-sharp | 0.24 | 0.00 | 397 | 0.14 | 0.33 |
| Opensmile (Male) | | | | | |
| Number of loudness peaks per second, indicating dynamic changes | -0.34 | 0.00 | 397 | -0.42 | -0.25 |
| Mean log amplitude of the 3rd formant (F3) relative to fundamental frequency | -0.32 | 0.00 | 397 | -0.40 | -0.22 |
| Mean log amplitude of the 1st formant (F1) relative to fundamental frequency | -0.32 | 0.00 | 397 | -0.41 | -0.23 |
| Mean log amplitude of the 1st formant (F1) relative to fundamental frequency | -0.31 | 0.00 | 397 | -0.40 | -0.22 |
| Median loudness distribution | -0.29 | 0.00 | 397 | -0.38 | -0.20 |
| Higher loudness levels | -0.28 | 0.00 | 397 | -0.37 | -0.18 |
| Range between 0th and 2nd percentiles of loudness | -0.28 | 0.00 | 397 | -0.37 | -0.19 |
| Mean loudness | -0.27 | 0.00 | 397 | -0.36 | -0.18 |
| Equivalent continuous sound level in dB, representing overall loudness | -0.24 | 0.00 | 397 | -0.33 | -0.14 |
| Standard deviation of voiced segment lengths in seconds | -0.23 | 0.00 | 397 | -0.32 | -0.13 |
| Overall mean spectral flux | -0.23 | 0.00 | 397 | -0.32 | -0.14 |
| Average duration of voiced segments in seconds | -0.21 | 0.00 | 397 | -0.30 | -0.11 |
| Mean slope of increasing loudness segments | -0.18 | 0.00 | 397 | -0.27 | -0.08 |
| Mean spectral flux in voiced regions | -0.16 | 0.00 | 397 | -0.26 | -0.07 |
| Mean ratio of spectral energy between 50-1000 Hz and 1-5 kHz in unvoiced regions | -0.15 | 0.01 | 397 | -0.24 | -0.05 |
| Mean slope of decreasing loudness segments | -0.15 | 0.01 | 397 | -0.25 | -0.06 |
| Lower pitch range | -0.14 | 0.02 | 397 | -0.23 | -0.04 |
| Mean spectral slope from 500-1500 Hz in voiced regions | 0.13 | 0.03 | 397 | 0.03 | 0.22 |
| Mean Hammarberg index in unvoiced regions, indicating spectral tilt | 0.14 | 0.02 | 397 | 0.04 | 0.23 |
| Mean local shimmer in dB, representing amplitude variation between cycles | 0.14 | 0.02 | 397 | 0.04 | 0.23 |
| Mean of finer spectral variation across all region | 0.16 | 0.00 | 397 | 0.07 | 0.26 |
| Average duration of unvoiced segments | 0.16 | 0.00 | 397 | 0.06 | 0.26 |
| Standard deviation of unvoiced segment lengths | 0.17 | 0.00 | 397 | 0.07 | 0.26 |
| Normalized standard deviation of loudness | 0.19 | 0.00 | 397 | 0.09 | 0.28 |
| Normalized standard deviation of spectral flux in voiced regions | 0.22 | 0.00 | 397 | 0.13 | 0.32 |
| Normalized standard deviation of overall spectral flux | 0.27 | 0.00 | 397 | 0.18 | 0.36 |
| Normalized standard deviation of F2 amplitude relative to fundamental frequency | 0.31 | 0.00 | 397 | 0.22 | 0.39 |
| Normalized standard deviation of F3 amplitude relative to fundamental frequency | 0.31 | 0.00 | 397 | 0.21 | 0.39 |
| Normalized standard deviation of F1 amplitude relative to fundamental frequency | 0.32 | 0.00 | 397 | 0.23 | 0.41 |
| Librosa (Female) | | | | | |
| Contrast in the third frequency sub-band | -0.24 | 0.00 | 397 | -0.33 | -0.15 |
| Contrast in the second frequency sub-band | -0.22 | 0.00 | 397 | -0.31 | -0.12 |
| Represents the overall spectral slope of the audio signal | -0.19 | 0.00 | 397 | -0.28 | -0.09 |
| Contrast in the fourth frequency sub-band | -0.18 | 0.00 | 397 | -0.27 | -0.08 |
| Contrast in the fifth frequency sub-band | -0.16 | 0.00 | 397 | -0.25 | -0.06 |
| Strength of the minor sixth interval | -0.13 | 0.02 | 397 | -0.22 | -0.03 |
| Strength of the perfect fifth interval | -0.13 | 0.02 | 397 | -0.22 | -0.03 |
| Higher-frequency spectral nuances | 0.12 | 0.03 | 397 | 0.02 | 0.22 |
| Fine spectral details, useful in music analysis | 0.13 | 0.02 | 397 | 0.03 | 0.22 |
| Intricate spectral patterns, useful for speaker identification | 0.15 | 0.01 | 397 | 0.05 | 0.25 |
| Energy in pitch class A | 0.15 | 0.01 | 397 | 0.05 | 0.24 |
| Energy in pitch class G-sharp | 0.16 | 0.00 | 397 | 0.06 | 0.26 |
| Energy in pitch class G | 0.19 | 0.00 | 397 | 0.10 | 0.29 |
| Energy in pitch class A-sharp | 0.19 | 0.00 | 397 | 0.09 | 0.28 |
| Energy in pitch class B | 0.20 | 0.00 | 397 | 0.10 | 0.29 |
| Energy in pitch class C | 0.21 | 0.00 | 397 | 0.11 | 0.30 |
| Energy in pitch class F-sharp | 0.24 | 0.00 | 397 | 0.15 | 0.33 |
| Energy in pitch class C-sharp | 0.26 | 0.00 | 397 | 0.16 | 0.35 |
| Energy in pitch class D | 0.27 | 0.00 | 397 | 0.18 | 0.36 |
| Energy in pitch class D-sharp | 0.28 | 0.00 | 397 | 0.19 | 0.37 |
| Energy in pitch class F | 0.29 | 0.00 | 397 | 0.20 | 0.38 |
| Energy in pitch class E | 0.29 | 0.00 | 397 | 0.19 | 0.37 |
| Opensmile (Female) | | | | | |
| Number of loudness peaks per second, indicating dynamic changes | -0.27 | 0.00 | 397 | -0.36 | -0.18 |
| Mean log amplitude of the 1st formant (F1) relative to fundamental frequency | -0.25 | 0.00 | 397 | -0.34 | -0.15 |
| Mean log amplitude of the 3rd formant (F3) relative to fundamental frequency | -0.25 | 0.00 | 397 | -0.34 | -0.16 |
| Mean log amplitude of the 1st formant (F1) relative to fundamental frequency | -0.25 | 0.00 | 397 | -0.34 | -0.16 |
| Median loudness distribution | -0.18 | 0.00 | 397 | -0.27 | -0.08 |
| Higher loudness levels | -0.18 | 0.00 | 397 | -0.27 | -0.08 |
| Overall mean spectral flux | -0.18 | 0.00 | 397 | -0.27 | -0.08 |
| Range between 0th and 2nd percentiles of loudness | -0.18 | 0.00 | 397 | -0.28 | -0.09 |
| Mean loudness | -0.17 | 0.01 | 397 | -0.26 | -0.07 |
| Mean log difference between fundamental frequency and amplitude of second harmonic | -0.13 | 0.05 | 397 | -0.22 | -0.03 |
| Higher pitch range | -0.13 | 0.04 | 397 | -0.23 | -0.03 |
| Mean spectral flux in unvoiced regions | -0.13 | 0.04 | 397 | -0.23 | -0.03 |
| Equivalent continuous sound level in dB, representing overall loudness | -0.13 | 0.04 | 397 | -0.23 | -0.03 |
| Normalized standard deviation of F3 frequency | 0.13 | 0.05 | 397 | 0.03 | 0.22 |
| Average duration of unvoiced segments | 0.15 | 0.02 | 397 | 0.05 | 0.24 |
| Normalized standard deviation of loudness | 0.16 | 0.01 | 397 | 0.06 | 0.25 |
| Normalized standard deviation of local shimmer | 0.16 | 0.01 | 397 | 0.06 | 0.25 |
| Normalized standard deviation of overall spectral flux | 0.17 | 0.00 | 397 | 0.07 | 0.26 |
| Mean bandwidth of F1 | 0.18 | 0.00 | 397 | 0.08 | 0.27 |
| Normalized standard deviation of F2 amplitude relative to fundamental frequency | 0.22 | 0.00 | 397 | 0.13 | 0.32 |
| Normalized standard deviation of F1 amplitude relative to fundamental frequency | 0.22 | 0.00 | 397 | 0.13 | 0.31 |
| Normalized standard deviation of F3 amplitude relative to fundamental frequency | 0.22 | 0.00 | 397 | 0.13 | 0.31 |

**Table S9: Audio Feature Correlations with Loneliness by Race**
Negative correlations indicate a higher correlation with low loneliness, positive indicates higher correlation with high loneliness ($p \leq 0.05$).

| Features | Pearson r | p-value | N | 95% CI (lower) | 95% CI (upper) |
|---|---|---|---|---|---|
| Librosa (White) | | | | | |
| Energy in pitch class C | 0.13 | 0.02 | 383 | 0.03 | 0.23 |
| Energy in pitch class C-sharp | 0.13 | 0.02 | 383 | 0.03 | 0.23 |
| Vocal tract shape | 0.15 | 0.01 | 383 | 0.05 | 0.25 |
| Higher-order spectral variations | 0.15 | 0.01 | 383 | 0.05 | 0.25 |
| Subtle difference in speech articulation | 0.15 | 0.01 | 383 | 0.05 | 0.25 |
| Finer spectral details beyond the primary formants | 0.16 | 0.00 | 383 | 0.06 | 0.26 |
| Energy in pitch class F-sharp | 0.16 | 0.00 | 383 | 0.06 | 0.26 |
| Energy in pitch class G | 0.16 | 0.00 | 383 | 0.06 | 0.26 |
| Energy in pitch class G-sharp | 0.17 | 0.00 | 383 | 0.07 | 0.26 |
| Energy in pitch class B | 0.20 | 0.00 | 383 | 0.10 | 0.30 |
| Fine spectral details, useful in music analysis | 0.21 | 0.00 | 383 | 0.11 | 0.30 |
| Energy in pitch class A | 0.22 | 0.00 | 383 | 0.12 | 0.31 |
| Higher-frequency spectral nuances | 0.24 | 0.00 | 383 | 0.14 | 0.33 |
| Energy in pitch class A-sharp | 0.26 | 0.00 | 383 | 0.17 | 0.35 |
| Captures intricate spectral patterns, useful for speaker identification | 0.27 | 0.00 | 383 | 0.17 | 0.36 |
| Opensmile (White) | | | | | |
| Mean log amplitude of the 3rd formant (F3) relative to fundamental frequency | -0.33 | 0.00 | 383 | -0.41 | -0.24 |
| Mean log amplitude of the 1st formant (F1) relative to fundamental frequency | -0.33 | 0.00 | 383 | -0.42 | -0.24 |
| Median loudness distribution | -0.32 | 0.00 | 383 | -0.41 | -0.22 |
| Mean log amplitude of the 1st formant (F1) relative to fundamental frequency | -0.32 | 0.00 | 383 | -0.41 | -0.23 |
| Number of loudness peaks per second, indicating dynamic changes | -0.31 | 0.00 | 383 | -0.40 | -0.21 |
| Mean loudness | -0.22 | 0.00 | 383 | -0.31 | -0.12 |

| Standard deviation of voiced segment lengths in seconds | -0.22 | 0.00 | 383 | -0.31 | -0.12 |
|---|---|---|---|---|---|
| Range between 0th and 2nd percentiles of loudness | -0.22 | 0.00 | 383 | -0.31 | -0.12 |
| Higher loudness levels | -0.22 | 0.00 | 383 | -0.32 | -0.13 |
| Overall mean spectral flux | -0.17 | 0.00 | 383 | -0.27 | -0.08 |
| Average duration of voiced segments in seconds | -0.16 | 0.01 | 383 | -0.26 | -0.06 |
| Equivalent continuous sound level in dB, representing overall loudness | -0.15 | 0.02 | 383 | -0.24 | -0.05 |
| Normalized standard deviation of F2 frequency | -0.13 | 0.05 | 383 | -0.23 | -0.03 |
| Normalized standard deviation of spectral slope from 0-500 Hz in voiced regions | 0.13 | 0.04 | 383 | 0.03 | 0.23 |
| Normalized standard deviation of spectral flux in voiced regions | 0.15 | 0.01 | 383 | 0.05 | 0.25 |
| Mean of finer spectral variation across all region | 0.16 | 0.01 | 383 | 0.06 | 0.26 |
| Standard deviation of unvoiced segment lengths | 0.19 | 0.00 | 383 | 0.09 | 0.29 |
| Average duration of unvoiced segments | 0.19 | 0.00 | 383 | 0.09 | 0.29 |
| Normalized standard deviation of loudness | 0.25 | 0.00 | 383 | 0.15 | 0.34 |
| Normalized standard deviation of overall spectral flux | 0.28 | 0.00 | 383 | 0.19 | 0.37 |
| Normalized standard deviation of F3 amplitude relative to fundamental frequency | 0.33 | 0.00 | 383 | 0.24 | 0.42 |
| Normalized standard deviation of F2 amplitude relative to fundamental frequency | 0.33 | 0.00 | 383 | 0.23 | 0.41 |
| Normalized standard deviation of F3 amplitude relative to fundamental frequency | 0.33 | 0.00 | 383 | 0.24 | 0.42 |
| Normalized standard deviation of F2 amplitude relative to fundamental frequency | 0.33 | 0.00 | 383 | 0.23 | 0.41 |
| Normalized standard deviation of F1 amplitude relative to fundamental frequency | 0.35 | 0.00 | 383 | 0.25 | 0.43 |
| Normalized standard deviation of F1 amplitude relative to fundamental frequency | 0.35 | 0.00 | 383 | 0.25 | 0.43 |
| Librosa (Black) | | | | | |
| Minute spectral features, aiding in emotion detection | 0.11 | 0.04 | 383 | 0.01 | 0.21 |
| Energy in pitch class B | 0.12 | 0.03 | 383 | 0.02 | 0.22 |
| Energy in pitch class C | 0.12 | 0.04 | 383 | 0.02 | 0.21 |
| Contrast in the highest frequency sub-band | 0.13 | 0.02 | 383 | 0.03 | 0.23 |
| Finer spectral details beyond the primary formants | 0.13 | 0.02 | 383 | 0.03 | 0.23 |

| Energy in pitch class D | 0.13 | 0.02 | 383 | 0.03 | 0.23 |
|---|---|---|---|---|---|
| Energy in pitch class D-sharp | 0.14 | 0.01 | 383 | 0.04 | 0.24 |
| Energy in pitch class C-sharp | 0.14 | 0.01 | 383 | 0.04 | 0.24 |
| Energy in pitch class G | 0.15 | 0.01 | 383 | 0.05 | 0.24 |
| Energy in pitch class F-sharp | 0.17 | 0.00 | 383 | 0.07 | 0.26 |
| Energy in pitch class G-sharp | 0.17 | 0.00 | 383 | 0.07 | 0.26 |
| Energy in pitch class E | 0.18 | 0.00 | 383 | 0.08 | 0.28 |
| Energy in pitch class A | 0.18 | 0.00 | 383 | 0.08 | 0.28 |
| Energy in pitch class A-sharp | 0.18 | 0.00 | 383 | 0.08 | 0.27 |
| Energy in pitch class F | 0.19 | 0.00 | 383 | 0.09 | 0.29 |
| Vocal tract shape | 0.20 | 0.00 | 383 | 0.10 | 0.29 |
| Captures intricate spectral patterns, useful for speaker identification | 0.24 | 0.00 | 383 | 0.14 | 0.33 |
| Higher-frequency spectral nuances | 0.24 | 0.00 | 383 | 0.14 | 0.33 |
| Fine spectral details, useful in music analysis | 0.27 | 0.00 | 383 | 0.17 | 0.36 |
| Opensmile (Black) | | | | | |
| Number of loudness peaks per second, indicating dynamic changes | -0.28 | 0.00 | 383 | -0.37 | -0.19 |
| Median loudness distribution | -0.27 | 0.00 | 383 | -0.36 | -0.17 |
| Mean log amplitude of the 3rd formant (F3) relative to fundamental frequency | -0.23 | 0.00 | 383 | -0.32 | -0.13 |
| Mean log amplitude of the 1st formant (F1) relative to fundamental frequency | -0.22 | 0.00 | 383 | -0.31 | -0.12 |
| Mean loudness | -0.20 | 0.00 | 383 | -0.29 | -0.10 |
| Range between 0th and 2nd percentiles of loudness | -0.20 | 0.00 | 383 | -0.29 | -0.10 |
| Mean log amplitude of the 1st formant (F1) relative to fundamental frequency | -0.20 | 0.00 | 383 | -0.30 | -0.10 |
| Higher loudness levels | -0.20 | 0.00 | 383 | -0.30 | -0.11 |
| Mean spectral slope from 0-500 Hz in unvoiced regions | -0.19 | 0.00 | 383 | -0.28 | -0.09 |
| Mean spectral flux in unvoiced regions | -0.18 | 0.00 | 383 | -0.28 | -0.09 |
| Higher pitch range | -0.17 | 0.00 | 383 | -0.26 | -0.07 |
| Normalized standard deviation of vocal tracts in voiced regions | -0.15 | 0.01 | 383 | -0.24 | -0.05 |
| Mean fundamental frequency in semitones above 27.5 Hz | -0.15 | 0.01 | 383 | -0.25 | -0.05 |
| Overall mean spectral flux | -0.15 | 0.01 | 383 | -0.25 | -0.05 |
| Median of fundamental frequency distribution | -0.14 | 0.02 | 383 | -0.24 | -0.04 |
| Mean log difference between fundamental frequency and amplitude of second harmonic | -0.14 | 0.01 | 383 | -0.24 | -0.05 |
| Mean frequency of F1 | -0.13 | 0.04 | 383 | -0.22 | -0.03 |
| Normalized standard deviation of local shimmer | 0.12 | 0.04 | 383 | 0.02 | 0.22 |
| Normalized standard deviation of lowest resonance frequency. | 0.12 | 0.04 | 383 | 0.02 | 0.22 |
| Mean spectral slope from 500-1500 Hz in unvoiced regions | 0.14 | 0.02 | 383 | 0.04 | 0.23 |
| Mean of vowel sounds in all regions | 0.15 | 0.01 | 383 | 0.05 | 0.25 |
| Mean of vowel sounds in voiced regions | 0.16 | 0.00 | 383 | 0.06 | 0.26 |
| Normalized standard deviation of overall spectral flux | 0.17 | 0.00 | 383 | 0.07 | 0.27 |
| Mean Hammarberg index in voiced regions | 0.17 | 0.00 | 383 | 0.07 | 0.27 |
| Normalized standard deviation of F3 bandwidth | 0.18 | 0.00 | 383 | 0.08 | 0.27 |
| Normalized standard deviation of loudness | 0.18 | 0.00 | 383 | 0.08 | 0.27 |
| Mean of spectral slope across voiced region | 0.19 | 0.00 | 383 | 0.09 | 0.29 |
| Normalized standard deviation of F1 amplitude relative to fundamental frequency | 0.19 | 0.00 | 383 | 0.09 | 0.29 |
| Mean of finer spectral variation across all region | 0.19 | 0.00 | 383 | 0.09 | 0.28 |
| Normalized standard deviation of local jitter | 0.19 | 0.00 | 383 | 0.09 | 0.28 |
| Normalized standard deviation of F2 frequency | 0.20 | 0.00 | 383 | 0.10 | 0.29 |
| Normalized standard deviation of F2 amplitude relative to fundamental frequency | 0.22 | 0.00 | 383 | 0.12 | 0.31 |
| Normalized standard deviation of F3 frequency | 0.23 | 0.00 | 383 | 0.13 | 0.32 |
| Normalized standard deviation of F3 amplitude relative to fundamental frequency | 0.23 | 0.00 | 383 | 0.13 | 0.32 |